\documentclass{article}

\usepackage{times}
\usepackage{graphicx} 
\usepackage{subfigure} 

\usepackage{natbib}

\usepackage{algorithm}
\usepackage{algorithmic}

\usepackage{amsmath}
\usepackage{amssymb}
\usepackage{amsfonts}
\usepackage{listings}

\newcommand{\Z}{\mathbb{Z}}
\newcommand{\R}{\mathbb{R}}

\newcommand{\corr}{\ensuremath{\star}}
\newcommand{\conv}{\ensuremath{*}}

\newcommand{\rpm}{\raisebox{.2ex}{$\scriptstyle\pm$}}
\newcommand{\rrpm}{\sbox0{$1$}\sbox2{$\scriptstyle\pm$}
  \raise\dimexpr(\ht0-\ht2)/2\relax\box2 }

\usepackage{hyperref}

\usepackage[accepted]{icml2016}

\icmltitlerunning{Group Equivariant Convolutional Networks}

\begin{document}

\twocolumn[
\icmltitle{Group Equivariant Convolutional Networks}

\icmlauthor{Taco S. Cohen}{t.s.cohen@uva.nl}
\icmladdress{University of Amsterdam}
\icmlauthor{Max Welling}{m.welling@uva.nl}
\icmladdress{
  University of Amsterdam \\
  University of California Irvine \\
  Canadian Institute for Advanced Research}

\icmlkeywords{}

\vskip 0.3in
]

\begin{abstract}
  We introduce Group equivariant Convolutional Neural Networks (G-CNNs), a natural generalization of convolutional neural networks that reduces sample complexity by exploiting symmetries.
  G-CNNs use G-convolutions, 
  a new type of layer that enjoys a substantially higher degree of weight sharing than regular convolution layers.
  G-convolutions increase the expressive capacity of the network without increasing the number of parameters.
  Group convolution layers are easy to use and can be implemented with negligible computational overhead for discrete groups generated by translations, reflections and rotations.
  G-CNNs achieve state of the art results on CIFAR10 and rotated MNIST.
\end{abstract}

\section{Introduction}
\label{sec:introduction}

Deep convolutional neural networks (CNNs, convnets) have proven to be very powerful models of sensory data such as images, video, and audio.
Although a strong theory of neural network design is currently lacking,
a large amount of empirical evidence supports the notion that both \emph{convolutional weight sharing} and \emph{depth} (among other factors) are important for good predictive performance.

Convolutional weight sharing is effective because there is a \emph{translation symmetry} in most perception tasks:
the label function and data distribution are both approximately invariant to shifts.
By using the same weights to analyze or model each part of the image,
a convolution layer uses far fewer parameters than a fully connected one, while preserving the capacity to learn many useful transformations.

Convolution layers can be used effectively in a \emph{deep} network because all the layers in such a network are \emph{translation equivariant}:
shifting the image and then feeding it through a number of layers is the same as feeding the original image through the same layers and then shifting the resulting feature maps (at least up to edge-effects).
In other words, the symmetry (translation) is preserved by each layer, which makes it possible to exploit it not just in the first, but also in higher layers of the network.

In this paper we show how convolutional networks can be generalized to exploit larger groups of symmetries, including rotations and reflections.
The notion of equivariance is key to this generalization,
so in section \ref{sec:equivariance} we will discuss this concept and its role in deep representation learning.
After discussing related work in section \ref{sec:related_work}, we recall a number of mathematical concepts in section \ref{sec:framework} that allow us to define and analyze the G-convolution in a generic manner. 

In section \ref{sec:translation_equivariance_of_CNNs}, we analyze the equivariance properties of standard CNNs, and show that they are equivariant to translations but may fail to equivary with more general transformations.
Using the mathematical framework from section \ref{sec:framework}, we can define G-CNNs (section \ref{sec:equivariance_of_GCNNs}) by analogy to standard CNNs (the latter being the G-CNN for the translation group).
We show that G-convolutions, as well as various kinds of layers used in modern CNNs, such as pooling, arbitrary pointwise nonlinearities, batch normalization and residual blocks are all equivariant, and thus compatible with G-CNNs.
In section \ref{sec:implementation} we provide concrete implementation details for group convolutions.

In section \ref{sec:experiments} we report experimental results on MNIST-rot and CIFAR10, where G-CNNs achieve state of the art results ($2.28\%$ error on MNIST-rot, and $4.19\%$ resp. $6.46\%$ on augmented and plain CIFAR10).
We show that replacing planar convolutions with G-convolutions consistently improves results without additional tuning.
In section \ref{sec:discussion} we provide a discussion of these results and consider several extensions of the method, before concluding in section \ref{sec:conclusion}.

\section{Structured \& Equivariant Representations}
\label{sec:equivariance}

Deep neural networks produce a sequence of progressively more abstract representations by mapping the input through a series of parameterized functions \cite{LeCun2015}.
In the current generation of neural networks, the representation spaces are usually endowed with very minimal internal structure, such as that of a linear space $\R^n$.

In this paper we construct representations that have the structure of a linear $G$-space, for some chosen group $G$.
This means that each vector in the representation space has a \emph{pose} associated with it, which can be transformed by the elements of some group of transformations $G$.
This additional structure allows us to model data more efficiently:
A filter in a G-CNN detects co-occurrences of features that have the preferred relative pose,
and can match such a feature constellation in every global pose through an operation called the G-convolution.

A representation space can obtain its structure from other representation spaces to which it is connected.
For this to work, the network or layer $\Phi$ that maps one representation to another should be \emph{structure preserving}. For $G$-spaces this means that $\Phi$ has to be equivariant:
\begin{equation}
  \label{eq:equivariance_definition}
  \Phi(T_g \, x) = T'_g \, \Phi(x),
\end{equation}
That is, transforming an input $x$ by a transformation $g$ (forming $T_g \, x$) and then passing it through the learned map $\Phi$ should give the same result as first mapping $x$ through $\Phi$ and then transforming the representation.

Equivariance can be realized in many ways, and in particular the operators $T$ and $T'$ need not be the same.
The only requirement for $T$ and $T'$ is that for any two transformations $g$ and $h$,
 we have $T(g h) = T(g) T(h)$ (i.e. $T$ is a \emph{linear representation} of $G$).

From equation \ref{eq:equivariance_definition} we see that the familiar concept of invariance is a special kind of equivariance where $T'_g$ is the identity transformation for all $g$.
In deep learning, general equivariance is more useful than invariance because it is impossible to determine if features are in the right spatial configuration if they are invariant.

Besides improving statistical efficiency and facilitating geometrical reasoning, equivariance to symmetry transformations constrains the network in a way that can aid generalization.
A network $\Phi$ can be non-injective, meaning that non-identical vectors $x$ and $y$ in the input space become identical in the output space (for example, two instances of a face may be mapped onto a single vector indicating the presence of any face).
If $\Phi$ is equivariant, then the $G$-transformed inputs $T_g \, x$ and $T_g \, y$ must also be mapped to the same output.
Their ``sameness'' (as judged by the network) is preserved under symmetry transformations.

\section{Related Work}
\label{sec:related_work}

There is a large body of literature on invariant representations.
Invariance can be achieved by pose normalization using an equivariant detector \cite{Lowe2004, Jaderberg2015} or by averaging a possibly nonlinear function over a group \cite{Reisert2008, Skibbe2013, Manay2006, Kondor2007}.

Scattering convolution networks use wavelet convolutions, nonlinearities and group averaging to produce stable invariants \cite{Bruna2013_TPAMI}.
Scattering networks have been extended to use convolutions on the group of translations, rotations and scalings, and have been applied to object and texture recognition \cite{Sifre2013, Oyallon2015}.

A number of recent works have addressed the problem of learning or constructing equivariant representations.
This includes work on transforming autoencoders \cite{Hinton2011}, equivariant Boltzmann machines \cite{Kivinen2011, Sohn2012}, equivariant descriptors \cite{Schmidt2012}, and equivariant filtering \cite{Skibbe2013}.

\citet{Lenc2014} show that the AlexNet CNN \cite{Krizhevsky2012} trained on imagenet spontaneously learns representations that are equivariant to flips, scaling and rotation.
This supports the idea that equivariance is a good inductive bias for deep convolutional networks.
\citet{Agrawal} show that useful representations can be learned in an unsupervised manner by training a convolutional network to be equivariant to ego-motion.

\citet{Anselmi2014, Anselmi2015} use the theory of locally compact topological groups to develop a theory of statistically efficient learning in sensory cortex.
This theory was implemented for the commutative group consisting of time- and vocal tract length shifts for an application to speech recognition by \citet{Zhang2015}.

\citet{Gensa} proposed an approximately equivariant convolutional architecture that uses sparse, high-dimensional feature maps to deal with high-dimensional groups of transformations.
\citet{Dieleman2015} showed that rotation symmetry can be exploited in convolutional networks for the problem of galaxy morphology prediction by rotating feature maps, effectively learning an equivariant representation.
This work was later extended \cite{Dieleman2016} and evaluated on various computer vision problems that have cyclic symmetry.

\citet{Cohen2014} showed that the concept of disentangling can be understood as a reduction of the operators $T_g$ in an equivariant representation, and later related this notion of disentangling to the more familiar statistical notion of decorrelation \cite{Cohen2015a}.

\section{Mathematical Framework}
\label{sec:framework}

In this section we present a mathematical framework that enables a simple and generic definition and analysis of G-CNNs for various groups $G$.
We begin by defining symmetry groups, and study in particular two groups that are used in the G-CNNs we have built so far.
Then we take a look at functions on groups (used to model feature maps in G-CNNs) and their transformation properties.

\subsection{Symmetry Groups}

A \emph{symmetry} of an object is a transformation that leaves the object invariant.
For example, if we take the sampling grid of our image, $\Z^2$, and flip it over we get $-\Z^2 = \{(-n, -m) \, | \, (n, m) \in \Z^2\} = \Z^2$.
So the flipping operation is a symmetry of the sampling grid.

If we have two symmetry transformations $g$ and $h$ and we compose them, the result $gh$ is another symmetry transformation (i.e. it leaves the object invariant as well).
Furthermore, the inverse $g^{-1}$ of any symmetry is also a symmetry, and composing it with $g$ gives the identity transformation $e$.
A set of transformations with these properties is called a \emph{symmetry group}.

One simple example of a group is the set of 2D integer translations, $\Z^2$.
Here the group operation (``composition of transformations'') is addition: $(n,m) + (p, q) = (n + p, m + q).$
One can verify that the sum of two translations is again a translation, and that the inverse (negative) of a translation is a translation, so this is indeed a group.

Although it may seem fancy to call 2-tuples of integers a group, this is helpful in our case because as we will see in section \ref{sec:equivariance_of_GCNNs},
a useful notion of convolution can be defined for functions on any group\footnote{At least, on any locally compact group.}, of which $\Z^2$ is only one example.
The important properties of the convolution, such as equivariance, arise primarily from the group structure.

\subsection{The group $p4$}
\label{sec:group_p4}

The group $p4$ consists of all compositions of translations and rotations by $90$ degrees about any center of rotation in a square grid.
A convenient parameterization of this group in terms of three integers $r, u, v$ is
\begin{equation}
  \label{eq:p4_matrix}
  g(r, u, v)
  =
  \begin{bmatrix}
    \cos\left(r \pi / 2 \right) & -\sin(r \pi / 2) & u \\
    \sin(r \pi / 2)             & \cos(r \pi / 2)  & v \\
    0                           & 0                & 1
  \end{bmatrix},
\end{equation}
where $0 \leq r < 4$ and $(u,v) \in \Z^2$.
The group operation is given by matrix multiplication.

The composition and inversion operations could also be represented directly in terms of integers $(r, u, v)$, but the equations are cumbersome.
Hence, our preferred method of composing two group elements represented by integer tuples is to convert them to matrices, multiply these matrices, and then convert the resulting matrix back to a tuple of integers (using the $\textup{atan2}$ function to obtain $r$).

The group $p4$ acts on points in $\Z^2$ (pixel coordinates) by multiplying the matrix $g(r, u, v)$ by the homogeneous coordinate vector $x(u', v')$ of a point $(u', v')$:
\begin{equation}
  gx
  \simeq
  \begin{bmatrix}
    \cos(r \pi / 2) & -\sin(r \pi / 2) & u \\
    \sin(r \pi / 2) & \cos(r \pi / 2) & v \\
    0 & 0 & 1
  \end{bmatrix}
  \begin{bmatrix}
    u' \\
    v' \\
    1
  \end{bmatrix}
\end{equation}

\subsection{The group $p4m$}
\label{sec:group_p4m}

The group $p4m$ consists of all compositions of translations, mirror reflections, and rotations by $90$ degrees about any center of rotation in the grid.
Like $p4$, we can parameterize this group by integers:
\begin{equation*}
  g(m, r, u, v) =
  \begin{bmatrix}
    (-1)^m \cos(\frac{r \pi}{2}) & -(-1)^{m} \sin(\frac{r \pi}{2}) & u \\
    \sin(\frac{r \pi}{2})        &  \cos(\frac{r \pi}{2})          & v \\
    0                            &  0                              & 1
  \end{bmatrix},
\end{equation*}
where $m \in \{0, 1\}$, $0 \leq r < 4$ and $(u,v)\in \Z^2$.
The reader may verify that this is indeed a group.

Again, composition is most easily performed using the matrix representation.
Computing $r, u, v$ from a given matrix $g$ can be done using the same method we use for $p4$, and for $m$ we have $m = \frac{1}{2} (1 - \det(g))$.

\subsection{Functions on groups}
\label{sec:feature_maps}

We model images and stacks of feature maps in a conventional CNN as functions $f : \Z^2 \rightarrow \R^K$ supported on a bounded (typically rectangular) domain.
At each pixel coordinate $(p, q) \in \Z^2$, the stack of feature maps returns a $K$-dimensional vector $f(p, q)$, where $K$ denotes the number of channels.

Although the feature maps must always be stored in finite arrays, modeling them as functions that extend to infinity (while being non-zero on a finite region only) simplifies the mathematical analysis of CNNs.

We will be concerned with transformations of the feature maps, so we introduce the following notation for a transformation $g$ acting on a set of feature maps:
\begin{equation}
  \label{eq:left_translation}
  [L_g f](x) = [f \circ g^{-1}](x) = f(g^{-1} x)
\end{equation}
Computationally, this says that to get the value of the $g$-transformed feature map $L_g f$ at the point $x$, we need to do a lookup in the original feature map $f$ at the point $g^{-1}x$, which is the unique point that gets mapped to $x$ by $g$.
This operator $L_g$ is a concrete instantiation of the transformation operator $T_g$ referenced in section \ref{sec:equivariance}, and one may verify that
\begin{equation}
  \label{eq:l_homomorphism}
  L_g L_h = L_{gh}.
\end{equation}

If $g$ represents a pure translation $t = (u,v) \in \Z^2$ then $g^{-1} x$ simply means $x - t$.
The inverse on $g$ in equation \ref{eq:left_translation} ensures that the function is shifted in the positive direction when using a positive translation, and that $L_g$ satisfies the criterion for being a homomorphism (eq. \ref{eq:l_homomorphism}) even for transformations $g$ and $h$ that do not commute (i.e. $gh \neq hg$).

As will be explained in section \ref{sec:g_equivariant_convolution}, feature maps in a $G$-CNN are functions on the group $G$, instead of functions on the group $\Z^2$.
For functions on $G$, the definition of $L_g$ is still valid if we simply replace $x$ (an element of $\Z^2$) by $h$ (an element of $G$), and interpret $g^{-1}h$ as composition.

\begin{figure}[!ht]
  \centering
  \includegraphics[width=0.20\textwidth]{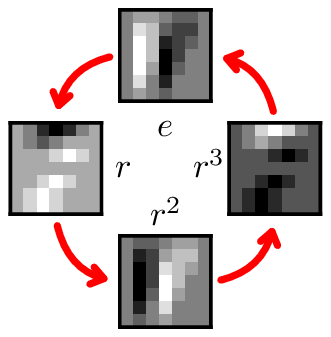}
  \includegraphics[width=0.20\textwidth]{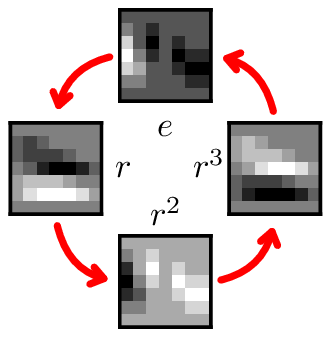}
  \caption{A p4 feature map and its rotation by $r$.}
  \label{fig:p4_fmap}
\end{figure}

It is easy to mentally visualize a planar feature map $f : \Z^2 \rightarrow \R$ undergoing a transformation, but we are not used to visualizing functions on groups.
To visualize a feature map or filter on $p4$, we plot the four patches associated with the four pure rotations on a circle, as shown in figure \ref{fig:p4_fmap} (left).
Each pixel in this figure has a rotation coordinate (the patch in which the pixel appears), and two translation coordinates (the pixel position within the patch).

When we apply the $90$ degree rotation $r$ to a function on $p4$, each planar patch follows its red $r$-arrow
(thus incrementing the rotation coordinate by $1$ (mod $4$)),
and simultaneously undergoes a $90$-degree rotation.
The result of this operation is shown on the right of figure \ref{fig:p4_fmap}.
As we will see in section \ref{sec:equivariance_of_GCNNs}, a $p4$ feature map in a $p4$-CNN undergoes exactly this motion under rotation of the input image.

\begin{figure}[!ht]
  \centering
  \includegraphics[width=0.20\textwidth]{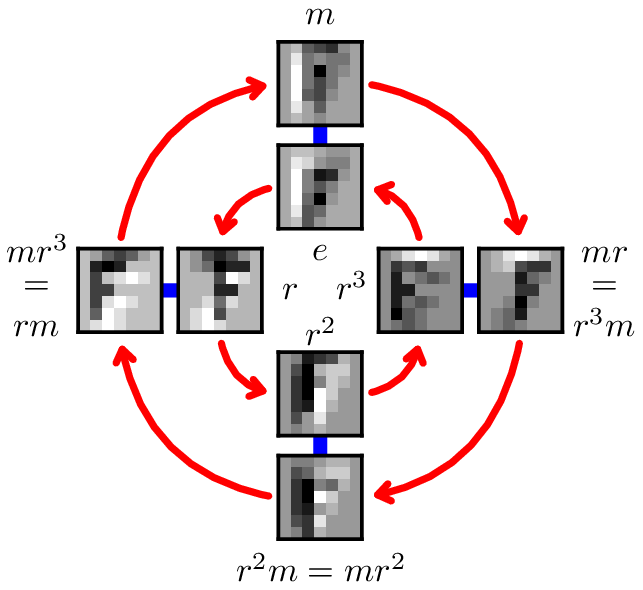}
  \includegraphics[width=0.20\textwidth]{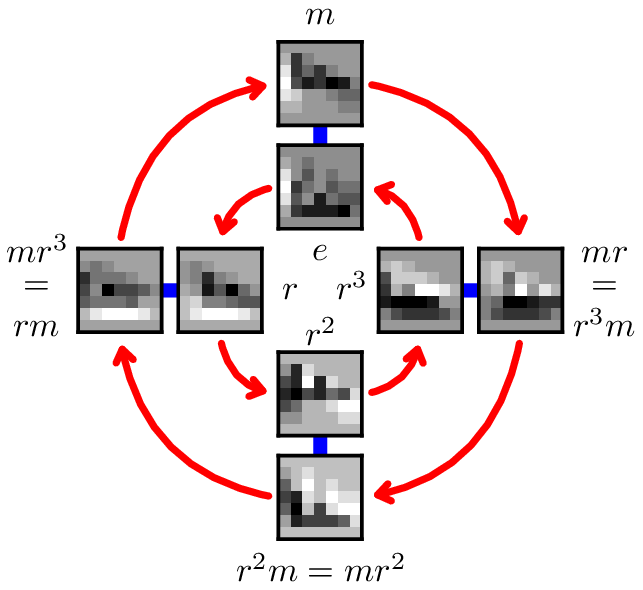}
  \caption{A p4m feature map and its rotation by $r$.}
  \label{fig:p4m_fmap}
\end{figure}

For $p4m$, we can make a similar plot, shown in figure \ref{fig:p4m_fmap}.
A $p4m$ function has $8$ planar patches, each one associated with a mirroring $m$ and rotation $r$.
Besides red rotation arrows, the figure now includes small blue reflection lines (which are undirected, since reflections are self-inverse).

Upon rotation of a $p4m$ function, each patch again follows its red $r$-arrows and undergoes a $90$ degree rotation.
Under a mirroring, the patches connected by a blue line will change places and undergo the mirroring transformation.
  
This rich transformation structure arises from the group operation of $p4$ or $p4m$, combined with equation \ref{eq:left_translation} which describes the transformation of a function on a group.

Finally, we define the \emph{involution} of a feature map, which will appear in section \ref{sec:g_equivariant_convolution} when we study the behavior of the G-convolution, and which also appears in the gradient of the G-convolution.
We have:
\begin{equation}
  \label{eq:involution}
  f^*(g) = f(g^{-1})
\end{equation}
For $\Z^2$ feature maps the involution is just a point reflection, but for $G$-feature maps the meaning depends on the structure of $G$.
In all cases, $f^{**} = f$.

\section{Equivariance properties of CNNs}
\label{sec:translation_equivariance_of_CNNs}

In this section we recall the definitions of the convolution and correlation operations used in conventional CNNs, and show that these operations are equivariant to translations but not to other transformations such as rotation.
This is certainly well known and easy to see by mental visualization, but deriving it explicitly will make it easier to follow the derivation of group equivariance of the group convolution defined in the next section. 

At each layer $l$, a regular convnet takes as input a stack of feature maps $f : \Z^2 \rightarrow \R^{K^l}$ and convolves or correlates it with a set of $K^{l+1}$ filters $\psi^i : \Z^2 \rightarrow \R^{K^l}$:
\begin{equation}
  \label{eq:conv_corr}
  \begin{aligned}
    \lbrack f \conv \psi^i](x) &= \sum_{y \in \Z^2} \sum_{k=1}^{K^l} f_k(y) \psi_k^i(x - y) \\
    [f \corr \psi^i](x) &= \sum_{y \in \Z^2} \sum_{k=1}^{K^l} f_k(y) \psi_k^i(y - x) \\
  \end{aligned}
\end{equation}

If one employs convolution ($\conv$) in the forward pass, the correlation ($\corr$) will appear in the backward pass when computing gradients, and vice versa.
We will use the correlation in the forward pass, and refer generically to both operations as ``convolution''.

Using the substitution $y \rightarrow y + t$, and leaving out the summation over feature maps for clarity, we see that a translation followed by a correlation is the same as a correlation followed by a translation:
\begin{equation}
  \label{eq:conv_equivariance}
  \begin{aligned}
    \lbrack[L_t f] \corr \psi](x)
    &=
    \sum_y f(y - t) \psi(y - x) \\
    &=
    \sum_y f(y) \psi(y + t - x) \\
    &=
    \sum_y f(y) \psi(y - (x - t)) \\
    &=
    [L_t [f \corr \psi]](x).
  \end{aligned}
\end{equation}
And so we say that ``correlation is an equivariant map for the translation group'', or that ``correlation and translation commute''.
Using an analogous computation one can show that also for the convolution, $[L_t f] \conv \psi = L_t [f \conv \psi]$.

Although convolutions are equivariant to \emph{translation}, they are not equivariant to other isometries of the sampling lattice.
For instance, as shown in the supplementary material, rotating the image and then convolving with a fixed filter is not the same as first convolving and then rotating the result:
\begin{equation}
  \begin{aligned}
    \lbrack [L_r f] \corr \psi](x)
      =
      L_r [f \corr [L_{r^{-1}} \psi]](x)
  \end{aligned}
\end{equation}
In words, this says that the correlation of a rotated image $L_r f$ with a filter $\psi$ is the same as the rotation by $r$ of the original image $f$ convolved with the inverse-rotated filter $L_{r^{-1}} \psi$.
Hence, if an ordinary CNN learns rotated copies of the same filter, the \emph{stack} of feature maps is equivariant, although individual feature maps are not.

\section{Group Equivariant Networks}
\label{sec:equivariance_of_GCNNs}

In this section we will define the three layers used in a G-CNN ($G$-convolution, $G$-pooling, nonlinearity) and show that each one commutes with $G$-transformations of the domain of the image.

\subsection{G-Equivariant correlation}
\label{sec:g_equivariant_convolution}

The correlation (eq. \ref{eq:conv_corr}) is computed by \emph{shifting} a filter and then computing a dot product with the feature maps.
By replacing the shift by a more general transformation from some group $G$, we get the $G$-correlation used in the first layer of a $G$-CNN:
\begin{equation}
  \label{eq:gconv_invariant}
    \lbrack f \corr \psi](g)
    =
    \sum_{y \in \Z^2} \sum_k f_k(y) \psi_k(g^{-1} y).
\end{equation}
Notice that both the input image $f$ and the filter $\psi$ are functions of the plane $\Z^2$, but the feature map $f \corr \psi$ is a function on the discrete group $G$ (which may contain translations as a subgroup).
Hence, for all layers after the first, the filters $\psi$ must also be functions on $G$, and the correlation operation becomes
\begin{equation}
  \label{eq:gconv}
    \lbrack f \corr \psi](g)
    =
    \sum_{h \in G} \sum_k f_k(h) \psi_k(g^{-1} h).  
\end{equation}

The equivariance of this operation is derived in complete analogy to eq. \ref{eq:conv_equivariance}, now using the substitution $h \rightarrow uh$:
\begin{equation}
  \label{eq:gconv_equivariance}
  \begin{aligned}
    \lbrack[L_u f] \corr \psi](g)
      &=
      \sum_{h \in G} \sum_k f_k(u^{-1} h) \psi(g^{-1} h) \\
      &=
      \sum_{h \in G} \sum_k f(h) \psi(g^{-1} u h) \\
      &=
      \sum_{h \in G} \sum_k f(h) \psi((u^{-1} g)^{-1} h) \\
      &=
      [L_u [f \corr \psi]](g)
  \end{aligned}
\end{equation}
The equivariance of eq. \ref{eq:gconv_invariant} is derived similarly.
Note that although equivariance is expressed by the same formula $[L_u f] \corr \psi = L_u [f \corr \psi]$ for both first-layer G-correlation (eq. \ref{eq:gconv_invariant}) and full G-correlation (\ref{eq:gconv}), the meaning of the operator $L_u$ is different:
for the first layer correlation, the inputs $f$ and $\psi$ are functions on $\Z^2$, so $L_u f$ denotes the transformation of such a function, while $L_u [f \corr \psi]$ denotes the transformation of the feature map, which is a function on $G$.
For the full $G$-correlation, both the inputs $f$ and $\psi$ and the output $f \corr \psi$ are functions on $G$.

Note that if $G$ is not commutative, neither the $G$-convolution nor the $G$-correlation is commutative.
However, the feature maps $\psi \corr f$ and $f \corr \psi$ are related by the involution (eq. \ref{eq:involution}):
\begin{equation}
  f \corr \psi = (\psi \corr f)^*.
\end{equation}
Since the involution is invertible (it is its own inverse), the information content of $f \corr \psi$ and $\psi \corr f$ is the same.
However, $f \corr \psi$ is more efficient to compute when using the method described in section \ref{sec:implementation}, because transforming a small filter is faster than transforming a large feature map.

It is customary to add a bias term to each feature map in a convolution layer.
This can be done for $G$-conv layers as well, as long as there is only one bias per $G$-feature map (instead of one bias per spatial feature plane within a $G$-feature map).
Similarly, batch normalization \cite{Ioffe2015} should be implemented with a single scale and bias parameter per $G$-feature map in order to preserve equivariance.
The sum of two $G$-equivariant feature maps is also $G$-equivariant, thus $G$-conv layers can be used in highway networks and residual networks \cite{Srivastava2015a, He2015}.

\subsection{Pointwise nonlinearities}

Equation \ref{eq:gconv_equivariance} shows that $G$-correlation preserves the transformation properties of the previous layer.
What about nonlinearities and pooling?

Recall that we think of feature maps as functions on $G$.
In this view, applying a nonlinearity $\nu : \R \rightarrow \R$ to a feature map amounts to \emph{function composition}.
We introduce the composition operator
\begin{equation}
  C_\nu f(g) = [\nu \circ f](g) = \nu(f(g)).
\end{equation}
which acts on functions by \emph{post}-composing them with $\nu$.

Since the left transformation operator $L$ acts by \emph{pre}-composition, $C$ and $L$ commute:
\begin{equation}
  C_\nu L_h f = \nu \circ [f \circ h^{-1}] = [\nu \circ f] \circ h^{-1} = L_h C_\nu f,
\end{equation}
so the rectified feature map inherits the transformation properties of the previous layer.

\subsection{Subgroup pooling and coset pooling}

In order to simplify the analysis, we split the pooling operation into two steps: the pooling itself (performed without stride), and a subsampling step.
The non-strided max-pooling operation applied to a feature map $f : G \rightarrow \R$ can be modeled as an operator $P$ that acts on $f$ as
\begin{equation}
  P f(g) = \max_{k \in g U} f(k),
\end{equation}
where $g U = \{gu \, | \, u \in U\}$ is the $g$-transformation of some pooling domain $U \subset G$ (typically a neighborhood of the identity transformation).
In a regular convnet, $U$ is usually a $2 \times 2$ or $3 \times 3$ square including the origin $(0, 0)$, and $g$ is a translation.

As shown in the supplementary material, pooling commutes with $L_h$:
\begin{equation}
    P L_h = L_h P
\end{equation}

Since pooling tends to reduce the variation in a feature map, it makes sense to sub-sample the pooled feature map, or equivalently, to do a ``pooling with stride''.
In a G-CNN, the notion of ``stride'' is generalized by subsampling on a subgroup $H \subset G$.
That is, $H$ is a subset of $G$ that is itself a group (i.e. closed under multiplication and inverses).
The subsampled feature map is then equivariant to $H$ but not $G$.

In a standard convnet, pooling with stride $2$ is the same as pooling and then subsampling on $H = \{(2i, 2 j) \, | (i,j) \in \Z^2\}$ which is a subgroup of $G = \Z^2$.
For the $p4$-CNN, we may subsample on the subgroup $H$ containing all $4$ rotations, as well as shifts by multiples of $2$ pixels.

We can obtain full $G$-equivariance by choosing our pooling region $U$ to be a subgroup $H \subset G$.
The pooling domains $g H$ that result are called \emph{cosets} in group theory.
The cosets partition the group into non-overlapping regions.
The feature map that results from pooling over cosets is invariant to the \emph{right}-action of $H$, because the cosets are similarly invariant ($gh H = g H$).
Hence, we can arbitrarily choose one coset representative per coset to subsample on.
The feature map that results from coset pooling may be thought of as a function on the quotient space $G / H$,
in which two transformations are considered equivalent if they are related by a transformation in $H$.

As an example, in a $p4$ feature map, we can pool over all four rotations at each spatial position (the cosets of the subgroup $R$ of rotations around the origin).
The resulting feature map is a function on $\Z^2 \cong p4 / R$, i.e. it will transform in the same way as the input image.
Another example is given by a feature map on $\Z$, where we could pool over the cosets of the subgroup $n \Z$ of shifts by multiples of $n$.
This gives a feature map on $\Z / n \Z$, which has a cyclic transformation law under translations.

This concludes our analysis of $G$-CNNs.
Since all layer types are equivariant, we can freely stack them into deep networks and expect G-conv parameter sharing to be effective at arbitrary depth.

\section{Efficient Implementation}
\label{sec:implementation}

Computing the G-convolution for involves nothing more than indexing arithmetic and inner products, so it can be implemented straightforwardly.
Here we present the details for a G-convolution implementation that can leverage recent advances in fast computation of planar convolutions \cite{Mathieu2014, Vasilache2015, Lavin2015}.

A plane symmetry group $G$ is called \emph{split} if any transformation $g \in G$ can be decomposed into a translation $t \in \Z^2$ and a transformation $s$ in the stabilizer of the origin (i.e. $s$ leaves the origin invariant).
For the group $p4$, we can write $g = ts$ for $t$ a translation and $s$ a rotation about the origin, while $p4m$ splits into translations and rotation-flips.
Using this split of $G$ and the fact that $L_g L_h = L_{gh}$, we can rewrite the G-correlation (eq. \ref{eq:gconv_invariant} and \ref{eq:gconv}) as follows:
\begin{equation}
  \label{eq:gcorr_decomp}
  f \corr \psi(ts) = 
  \sum_{h \in X} \sum_k f_k(h) L_{t} \left[ L_s \psi_k(h) \right]
\end{equation}
where $X = \Z^2$ in layer one and $X = G$ in further layers.

Thus, to compute the $p4$ (or $p4m$) correlation $f \corr \psi$ we can first compute $L_s \psi$ (``filter transformation'') for all four rotations (or all eight rotation-flips) and then call a fast planar correlation routine on $f$ and the augmented filter bank.

The computational cost of the algorithm presented here is roughly equal to that of a planar convolution with a filter bank that is the same size as the augmented filter bank used in the G-convolution, because the cost of the filter transformation is negligible.

\subsection{Filter transformation}

The set of filters at layer $l$ is stored in an array $F[\cdot]$ of shape $K^l \times K^{l-1} \times S^{l-1} \times n \times n$, where $K^l$ is the number of channels at layer $l$, $S^{l-1}$ denotes the number of transformations in $G$ that leave the origin invariant (e.g. $1$, $4$ or $8$ for $\Z^2$, $p4$ or $p4m$ filters, respectively), and $n$ is the spatial (or translational) extent of the filter.
Note that typically, $S^1 = 1$ for 2D images, while $S^l = 4$ or $S^l = 8$ for $l>1$.

The filter transformation $L_s$ amounts to a permutation of the entries of each of the $K^l \times K^{l-1}$ scalar-valued filter channels in $F$.
Since we are applying $S^l$ transformations to each filter, the output of this operation is an array of shape $K^l \times S^l \times K^{l-1} \times S^{l-1} \times n \times n$, which we call $F^+$.

The permutation can be implemented efficiently by a GPU kernel that does a lookup into $F$ for each output cell of $F^+$, using a precomputed index associated with the output cell.
To precompute the indices, we define an invertible map $g(s, u, v)$ that takes an input index (valid for an array of shape $S^{l-1} \times n \times n$) and produces the associated group element $g$ as a matrix (section \ref{sec:group_p4} and \ref{sec:group_p4m}).
For each input index $(s, u, v)$ and each transformation $s'$, we compute $\bar{s}, \bar{u}, \bar{v} = g^{-1}(g(s', 0, 0)^{-1} g(s, u, v))$.
This index is used to set $F^+[i, s', j, s, u, v] = F[i, j, \bar{s}, \bar{u}, \bar{v}]$ for all $i,j$.

The G-convolution for a new group can be added by simply implementing a map $g(\cdot)$ from indices to matrices.

\subsection{Planar convolution}

The second part of the G-convolution algorithm is a planar convolution using the expanded filter bank $F^+$.
If $S^{l-1} > 1$, the sum over $X$ in eq. \ref{eq:gcorr_decomp} involves a sum over the stabilizer.
This sum can be folded into the sum over feature channels performed by the planar convolution routine by reshaping $F^+$ from $K^l \times S^l \times K^{l-1} \times S^{l-1} \times n \times n$ to $S^l K^l \times S^{l-1} K^{l-1} \times n \times n$.
The resulting array can be interpreted as a conventional filter bank with $S^{l-1} K^{l-1}$ planar input channels and $S^l K^l$ planar output channels, which can be correlated with the feature maps $f$ (similarly reshaped).

\section{Experiments}
\label{sec:experiments}

\subsection{Rotated MNIST}

The rotated MNIST dataset \cite{Larochelle2007} contains $62000$ randomly rotated handwritten digits.
The dataset is split into a training, validation and test sets of size $10000$, $2000$ and $50000$, respectively.

We performed model selection using the validation set, yielding a CNN architecture (Z2CNN) with $7$ layers of $3\times 3$ convolutions ($4 \times 4$ in the final layer), $20$ channels in each layer, relu activation functions, batch normalization, dropout, and max-pooling after layer $2$.
For optimization, we used the Adam algorithm \cite{Kingma2015}.
This baseline architecture outperforms the models tested by \citet{Larochelle2007} (when trained on $12$k and evaluated on $50$k), but does not match the previous state of the art, which uses prior knowledge about rotations \cite{Schmidt2012} (see table \ref{tbl:mnist_rot_results}).

Next, we replaced each convolution by a $p4$-convolution (eq. \ref{eq:gconv_invariant} and \ref{eq:gconv}), divided the number of filters by $\sqrt{4} = 2$ (so as to keep the number of parameters approximately fixed), and added max-pooling over rotations after the last convolution layer.
This architecture (P4CNN) was found to perform better without dropout, so we removed it.
The P4CNN almost halves the error rate of the previous state of the art ($2.28\%$ vs $3.98\%$ error).

We then tested the hypothesis that premature invariance is undesirable in a deep architecture (section \ref{sec:equivariance}).
We took the Z2CNN, replaced each convolution layer by a $p4$-convolution (eq. \ref{eq:gconv_invariant}) followed by a coset max-pooling over rotations.
The resulting feature maps consist of rotation-invariant features, and have the same transformation law as the input image.
This network (P4CNNRotationPooling) outperforms the baseline and the previous state of the art, but performs significantly worse than the P4CNN which does not pool over rotations in intermediate layers.

\begin{table}[h!]
  \centering
  \begin{tabular}{l r}
    \hline
    Network & Test Error (\%) \\
    \hline
    \citet{Larochelle2007} & 10.38 \rpm \, 0.27  \\
    \citet{Sohn2012}       & 4.2 \\
    \citet{Schmidt2012}    & 3.98  \\
    \hline
    Z2CNN                  & 5.03 \rpm \, 0.0020 \\
    P4CNNRotationPooling   & 3.21 \rpm \, 0.0012 \\
    \bf{P4CNN}             & \bf{2.28} \rpm \, 0.0004 \\
    \hline 
  \end{tabular}
  \caption{Error rates on rotated MNIST (with standard deviation under variation of the random seed).}
  \label{tbl:mnist_rot_results}
\end{table}

\subsection{CIFAR-10}

The CIFAR-10 dataset consists of $60k$ images of size $32 \times 32$, divided into $10$ classes.
The dataset is split into $40k$ training, $10k$ validation and $10k$ testing splits.

We compared the $p4$-, $p4m$- and standard planar $\Z^2$ convolutions on two kinds of baseline architectures.
Our first baseline is the All-CNN-C architecture by \citet{Springenberg2015},
which consists of a sequence of $9$ strided and non-strided convolution layers, interspersed with rectified linear activation units, and nothing else.
Our second baseline is a residual network \cite{He2016}, which consists of an initial convolution layer, followed by three stages of $2 n$ convolution layers using $k_i$ filters at stage $i$, followed by a final classification layer ($6n+2$ layers in total).
The first convolution in each stage $i > 1$ uses a stride of 2, so the feature map sizes are 32, 16, and 8 for the three stages.
We use $n = 7$, $k_i = 32, 64, 128$ yielding a wide $44$-layer network called ResNet44.

To evaluate G-CNNs, we replaced all convolution layers of the baseline architectures by $p4$ or $p4m$ convolutions.
For a constant number of filters, this increases the size of the feature maps $4$ or $8$-fold, which in turn increases the number of parameters required per filter in the next layer.
Hence, we halve the number of filters in each $p4$-conv layer, and divide it by roughly $\sqrt{8} \approx 3$ in each $p4m$-conv layer.
This way, the number of parameters is left approximately invariant, while the size of the internal representation is increased.
Specifically, we used $k_i = 11, 23, 45$ for $p4m$-ResNet44.

To evaluate the impact of data augmentation, we compare the networks on CIFAR10 and augmented CIFAR10+.
The latter denotes moderate data augmentation with horizontal flips and small translations, following \citet{Goodfellow2013} and many others.

The training procedure for training the All-CNN was reproduced as closely as possible from \citet{Springenberg2015}.
For the ResNets, we used stochastic gradient descent with initial learning rate of $0.05$ and momentum $0.9$.
The learning rate was divided by $10$ at epoch $50, 100$ and $150$, and training was continued for $300$ epochs.

\begin{table}[h!]
  \centering
  \begin{tabular}{l | c c c r}
    \hline
    Network   & $G$    & CIFAR10 & CIFAR10+ & Param. \\
    \hline
    All-CNN   & $\Z^2$  & 9.44 & 8.86 & 1.37M \\
              & $p4$    & 8.84 & 7.67 & 1.37M \\
              & $p4m$   & 7.59 & 7.04 & 1.22M \\
    ResNet44  & $\Z^2$  & 9.45 & 5.61 & 2.64M \\
              & $p4m$   & \bf{6.46} & \bf{4.94} & 2.62M \\
    \hline
  \end{tabular}
  \caption{Comparison of conventional (i.e. $\Z^2$), $p4$ and $p4m$ CNNs on CIFAR10 and augmented CIFAR10+. Test set error rates and number of parameters are reported.}
  \label{tbl:cifar10_results}
\end{table}

To the best of our knowledge, the $p4m$-CNN outperforms all published results on plain CIFAR10 \cite{Wan2013, Goodfellow2013, Lin2013, Lee2015, Srivastava2015a, Clevert2015, Lee2015a}.
However, due to radical differences in model sizes and architectures, it is difficult to infer much about the intrinsic merit of the various techniques.
It is quite possible that the cited methods would yield better results when deployed in larger networks or in combination with other techniques.
Extreme data augmentation and model ensembles can also further improve the numbers \cite{Graham}.

Inspired by the wide ResNets of \citet{Zagoruyko2016}, we trained another ResNet with $26$ layers and $k_i = (71, 142, 248)$ (for planar convolutions) or $k_i = (50, 100, 150)$ (for $p4m$ convolutions).
When trained with moderate data augmentation, this network achieves an error rate of $5.27\%$ using planar convolutions, and $\mathbf{4.19\%}$ with $p4m$ convolutions.
This result is comparable to the $4.17\%$ error reported by \citet{Zagoruyko2016}, but using fewer parameters (7.2M vs 36.5M).

\section{Discussion \& Future work}
\label{sec:discussion}

Our results show that $p4$ and $p4m$ convolution layers can be used as a drop-in replacement of standard convolutions that consistently improves the results.

G-CNNs benefit from data augmentation in the same way as convolutional networks,
as long as the augmentation comes from a group larger than $G$.
Augmenting with flips and small translations consistently improves the results for the $p4$ and $p4m$-CNN.

The CIFAR dataset is not actually symmetric, since objects typically appear upright.
Nevertheless, we see substantial increases in accuracy on this dataset, indicating that there need not be a full symmetry for G-convolutions to be beneficial.

In future work, we want to implement G-CNNs that work on hexagonal lattices which have an increased number of symmetries relative to square grids, as well as G-CNNs for 3D space groups.
All of the theory presented in this paper is directly applicable to these groups, and the G-convolution can be implemented in such a way that new groups can be added by simply specifying the group operation and a bijective map between the group and the set of indices.

One limitation of the method as presented here is that it only works for discrete groups.
Convolution on continuous (locally compact) groups is mathematically well-defined, but may be hard to approximate in an equivariant manner.
A further challenge, already identified by \citet{Gensa}, is that a full enumeration of transformations in a group may not be feasible if the group is large.

Finally, we hope that the current work can serve as a concrete example of the general philosophy of ``structured representations'', outlined in section \ref{sec:equivariance}.
We believe that adding mathematical structure to a representation (making sure that maps between representations preserve this structure), could enhance the ability of neural nets to see abstract similarities between superficially different concepts.

\section{Conclusion}
\label{sec:conclusion}
We have introduced G-CNNs, a generalization of convolutional networks that 
substantially increases the expressive capacity of a network without increasing the number of parameters.
By exploiting symmetries, G-CNNs achieve state of the art results on rotated MNIST and CIFAR10.
We have developed the general theory of G-CNNs for discrete groups, showing that all layer types are equivariant to the action of the chosen group $G$.
Our experimental results show that G-convolutions can be used as a drop-in replacement for spatial convolutions in modern network architectures, improving their performance without further tuning.

\section*{Acknowledgements}
We would like to thank Joan Bruna, Sander Dieleman, Robert Gens, Chris Olah, and Stefano Soatto for helpful discussions.
This research was supported by NWO (grant number NAI.14.108), Google and Facebook.

\bibliography{icml2016}
\bibliographystyle{icml2016}

\newpage
\section*{Appendix A: Equivariance Derivations}

We claim in the paper that planar correlation is not equivariant to rotations.
Let $f : \R^2 \rightarrow \R^K$ be an image with $K$ channels, and let $\psi : \R^2 \rightarrow \R^K$ be a filter.
Take a rotation $r$ about the origin.
The ordinary planar correlation $\star$ is not equivariant to rotations, i.e., $[L_r f] \star \psi \neq L_r [f \star \psi]$.
Instead we have:
\begin{equation}
  \begin{aligned}
    \lbrack [L_r f] \corr \psi](x)
      &=
      \sum_{y\in \Z^2} \sum_k L_r f_k(y) \psi_k(y - x) \\
      &=
      \sum_{y\in \Z^2} \sum_k f_k(r^{-1} y) \psi_k(y - x) \\
      &=
      \sum_{y\in \Z^2} \sum_k f_k(y) \psi_k(ry - x) \\
      &=
      \sum_{y\in \Z^2} \sum_k f_k(y) \psi_k(r(y - r^{-1}x)) \\
      &=
      \sum_{y\in \Z^2} \sum_k f_k(y) L_{r^{-1}} \psi(y - r^{-1} x)) \\
      &=
      f \corr [L_{r^{-1}} \psi](r^{-1} x) \\
      &=
      L_r [f \corr [L_{r^{-1}} \psi]](x)
  \end{aligned}
\end{equation}

Line by line, we used the following definitions, facts and manipulations:
\begin{enumerate}
  \item The definition of the correlation $\corr$.
  \item The definition of $L_r$, i.e. $L_r f(x) = f(r^{-1} x)$.
  \item The substitution $y \rightarrow ry$, which does not change the summation bounds since rotation is a symmetry of the sampling grid $\Z^2$.
  \item Distributivity.
  \item The definition of $L_r$.
  \item The definition of the correlation $\corr$.
  \item The definition of $L_r$.
\end{enumerate}

A visual proof can be found in \cite{Dieleman2016}.

Using a similar line of reasoning, we can show that pooling commutes with the group action:
\begin{equation}
  \begin{aligned}
    P L_h f(g)
    &=
    \max_{k \in gU} L_h f(k) \\
    &=
    \max_{k \in gU} f(h^{-1} k) \\
    &=
    \max_{hk \in g U} f(k) \\
    &=
    \max_{k \in h^{-1} g U} f(k) \\
    &=
    P f(h^{-1} g) \\
    &=
    L_h P f(g)
  \end{aligned}
\end{equation}

\section*{Appendix B: Gradients}

To train a G-CNN, we need to compute gradients of a loss function with respect to the parameters of the filters.
If we use the fast algorithm explained in section 7 of the main paper, we only have to implement the gradient of the indexing operation (section 7.1, ``filter transformation''), because the 2D convolution routine and its gradient are given.

This gradient is computed as follows.
The gradient of the loss with respect to cell $i$ in the input of the indexing operation is the sum of the gradients of the output cells $j$ that index cell $i$.
On current GPU hardware, this can be implemented efficiently using a kernel that is instantiated for each cell $j$ in the \emph{output} array.
The kernel adds the value of the gradient of the loss with respect to cell $j$ to cell $i$ of the array that holds the gradient of the loss with respect to the input of the indexing operation (this array is to be initialized at zero).
Since multiple kernels write to the same cell $i$, the additions must be done using atomic operations to avoid concurrency problems.

Alternatively, one could implement the filter transformation using a precomputed permutation matrix.
This is not as efficient, but the gradient is trivial, and most computation graph / deep learning packages will have implemented the matrix multiplication and its gradient.

\section*{Appendix C: G-conv calculus}

Although the gradient of the filter transformation operation is all that is needed to do backpropagation in a G-CNN for a split group $G$, it is instructive to derive the analytical gradients of the G-correlation operation.
This leads to an elegant ``G-conv calculus'', included here for the interested reader.

Let feature map $k$ at layer $l$ be denoted $f^l_k = f^{l-1} \corr \psi^{lk}$, where $f^{l-1}$ is the stack of feature maps in the previous layer.
At some point in the backprop algorithm, we will have computed the derivative $\partial L / \partial f^l_k$ for all $k$, and we need to compute $\partial L / \partial f^{l-1}_j$ (to backpropagate to lower layers) as well as $\partial L / \partial \psi_j^{lk}$ (to update the parameters).
We find that,
\begin{equation}
  \begin{aligned}
    \frac{\partial L}{\partial f_j^{l-1}(g)}
    &=
    \sum_{h,k} \frac{\partial L}{\partial f^l_k(h)} \frac{\partial f^l_k(h)}{\partial f^{l-1}_j(g)} \\
    &=
    \sum_{h,k} \frac{\partial L}{\partial f^l_k(h)} \left[\sum_{h', k'} \frac{\partial f^{l-1}_{k'}(h')}{\partial f^{l-1}_j(g)} \psi^{lk}_{k'}(h^{-1} h') \right] \\
    &=
    \sum_{h,k} \frac{\partial L}{\partial f^l_k(g)} \psi^{lk}_{j}(h^{-1} g) \\
    &=
    \left[ \frac{\partial L}{\partial f^l} \corr \psi^{l*}_j \right](g)
  \end{aligned}
\end{equation}
where the superscript $^*$ denotes the involution 
\begin{equation}
  \psi^*(g) = \psi(g^{-1}),
\end{equation}
and $\psi_j^l$ is the set of filter components applied to input feature map $j$ at layer $l$:
\begin{equation}
  \psi_j^l(g) = (\psi_j^{l1}(g), \ldots, \psi_j^{lK_l}(g))
\end{equation}

To compute the gradient with respect to component $j$ of filter $k$, we have to G-convolve the $j$-th input feature map with the $k$-th output feature map:
\begin{equation}
  \begin{aligned}
    \frac{\partial L}{\partial \psi_j^{lk}(g)}
    &=
    \sum_{h} \frac{\partial L}{\partial f^l_k(h)} \frac{\partial f^l_k(h)}{\partial \psi_j^{lk}(g)} \\
    &=
    \sum_{h} \frac{\partial L}{\partial f^l_k(h)} \left[\sum_{h', k'} f^{l-1}_{k'}(h') \frac{\partial \psi^{lk}_{k'}(h^{-1} h')}{\partial \psi_j^{lk}(g)} \right] \\
    &=
    \sum_{h} \frac{\partial L}{\partial f^l_k(h)} f^{l-1}_{j}(h g) \\
    &=
    \left[ \frac{\partial L}{\partial f^l_k} \conv f_j^{l-1} \right] (g)
  \end{aligned}
\end{equation}

So we see that both the forward and backward passes involve convolution or correlation operations, as is the case in standard convnets.

\end{document}